\PassOptionsToPackage{pdfpagelabels=false}{hyperref}
\documentclass{IOS-Book-Article}     
\usepackage[T1]{fontenc}
\usepackage[english]{babel}
\usepackage{cite}
\usepackage{graphicx}
\usepackage{xcolor}

\begin{document}
\begin{frontmatter}          
%

\title{A Neuro-Symbolic Multi-Agent Approach to Legal–Cybersecurity Knowledge Integration}



\runningtitle{}

%
\author[A]{\fnms{Chiara} \snm{Bonfanti}}
\author[B]{\fnms{Alessandro} \snm{Druetto}}
\author[A]{\fnms{Cataldo} \snm{Basile}}
\author[C]{\fnms{Tharindu} \snm{Ranasinghe}}
\author[D] {\fnms{Marcos} \snm{Zampieri}}
\address[A]{Department of Control and Computer Engineering, Politecnico di Torino, Italy}
\address[B]{Department of Computer Science, Università di Torino, Italy}
\address[C]{School of Computing and Communications, Lancaster University, UK}
\address[D]{School of Computing, George Mason University, USA}

\begin{abstract}

The growing intersection of cybersecurity and law creates a complex information space where traditional legal research tools struggle to deal with nuanced connections between cases, statutes, and technical vulnerabilities. This knowledge divide hinders collaboration between legal experts and cybersecurity professionals. To address this important gap, this work provides a first step towards intelligent systems capable of navigating the increasingly intricate cyber-legal domain. We demonstrate promising initial results on multilingual tasks.


\end{abstract}

\begin{keyword}
Cybersecurity, Legal Informatics, NLP, European Digital Law, Case Retrieval, Hybrid Architectures

\end{keyword}

\end{frontmatter}


\section*{Introduction}

As cyber threats become more sophisticated, a significant knowledge gap emerges between two essential domains: legal experts who lack a deep technical understanding of cybersecurity mechanisms, and technical professionals who are unfamiliar with legal doctrine and jurisprudential reasoning~\cite{BenchCapon2017, Jaiswal2024artificial}. This divide impedes compliance and raises operational risk. Automated approaches are therefore needed to bridge legal and technical domains in a way that is both reliable and interpretable. Existing methods remain limited: keyword-based retrieval often fails to capture the semantic links between legal norms and attack patterns, while neural models such as large language models (LLMs) provide richer semantics but lack transparency and may introduce errors through hallucination~\cite{Chalkidis2021}. Recent work has advanced the extraction of deontic obligations from EU regulations~\cite{RaulinoDalPont2025}, yet the technical dimensions of cybersecurity remain largely unaddressed.

We propose a neuro-symbolic multi-agent framework that explicitly links legal obligations to technical documentation, including, but not limited to, the NIST Special Publications and the MITRE resources, which are widely de facto standards in cybersecurity \cite{nist80053}. 
These mappings are integrated into a knowledge graph that supports retrieval and reasoning through reinforcement learning and Belief–Desire–Intention (BDI) agents \cite{bdi}. By combining knowledge graph reasoning with rule-based classification, the framework achieves a level of explainability and auditability often absent in purely neural approaches. To our knowledge, this work is among the first systems to make machine-readable the connections between legal obligations and cybersecurity concepts, enabling transparent and traceable reasoning for both legal practitioners and technical experts.

\section*{Background}
\textbf{The CEPS-ZENNER dataset.}
Within the landscape of European legal resources, the CEPS--Zenner dataset occupies a distinctive position~\cite{zenner2025dataset}. Earlier corpora, such as EUR-Lex and the JRC-Acquis~\cite{Steinberger2006}, and MetaLex~\cite{Agnoloni2008, Boer2010}, have primarily concentrated on legislative texts and multilingual access. At the same time, initiatives such as Akoma Ntoso~\cite{Palmirani2011}, the European Legislation Identifier (ELI)~\cite{VanOpijnen2011}, and LegalRuleML~\cite{Governatori2013} have focused on machine-readable representations of statutes. Within AI \& Law, a rich body of work has explored legal ontologies and structured knowledge representation~\cite{Boer2010, Francesconi2019}, argumentation-based reasoning~\cite{BenchCapon2017}, and normative multi-agent systems~\cite{Boella2007}, demonstrating the importance of modelling institutions and their rules. CEPS-Zenner systematically maps digital policy measures enacted, proposed, and planned with the governance bodies responsible for implementation. This institutional and temporal perspective is critical for domains such as cybersecurity, where regulatory obligations intersect with technical enforcement. While originally curated for policy analysis and lacking alignment with semantic standards, CEPS-Zenner provides a challenging and valuable substrate for multilingual, neuro-symbolic, multi-agent experiments. While CEPS-Zenner was curated for policy analysis rather than computational use, and lacks alignment with semantic standards such as Akoma Ntoso or LegalRuleML, these very limitations make it a challenging yet valuable field for experiments with our multilingual, neuro-symbolic multi-agent framework.

\textbf{Cybersecurity technical documentation.}
The cybersecurity roles that usually interface with legal informatics are Chief Information Security Officers, senior cybersecurity designers, and analysts. 
Legal frameworks, especially those that focus on compliance (e.g., the EU GDPR and NIS2), require that the cybersecurity risks are kept under a tolerable level. 
One fundamental step these roles must perform is to map the high-level legal constraints to the threats that can increase the risks and the mitigations that could lower them again to acceptable levels. 
To this purpose, the most relevant and structured results have been produced by the Mitre Corporation,  ATT\&CK\footnote{\url{https://attack.mitre.org/resources/attack-data-and-tools/}}, which describes adversary tactics and techniques, and D3FEND\footnote{\url{https://d3fend.mitre.org/resources/}}, which details defensive techniques and technologies to counter malicious cyber threats.

\textbf{AI and Automation in Legal Informatics.}
Integrating AI into legal informatics has seen significant advancements, particularly in automating tasks such as contract analysis, compliance monitoring, and case prediction. However, challenges persist in bridging the gap between legal reasoning and technical processes. Recent studies highlight the potential of AI agents to enhance legal workflows, though human oversight remains essential due to concerns over AI errors and ``hallucinations''~\cite{BenchCapon2017, Chalkidis2021}. This underscores the necessity for AI systems that are both effective and interpretable within legal contexts.

\textbf{Knowledge Representation and Reasoning.} Traditional knowledge representation in legal informatics has focused on formalizing legal norms using frameworks such as LegalRuleML and Akoma Ntoso, but these often lack integration with real-world data and dynamic reasoning capabilities~\cite{Governatori2013, Palmirani2011, Boer2010, Francesconi2019}. Recent approaches leverage reinforcement learning, e.g., information-gain-based rewards in LLMs, to enable more flexible and robust legal reasoning~\cite{LegalDelta2025}.

\textbf{Reinforcement Learning.}
RL has been applied to various aspects of legal information retrieval. For example, the Reinforced-IR framework employs a self-boosting mechanism to adapt pre-trained retrievers and generators for precise cross-domain retrieval~\cite{ReinforcedIR2025}. Additionally, beam search techniques have been utilized in legal case retrieval to efficiently generate relevant case identifiers, accommodating the hierarchical nature of legal structures~\cite{BeamSearchLegal2025}. These methods enhance the accuracy and efficiency of legal information retrieval systems.

\textbf{Beam Search.} Beam search~\cite{Lowerre1976} is a heuristic search algorithm widely used for approximate decoding in sequence generation tasks, such as machine translation~\cite{Freitag2017} and speech recognition.
It extends the greedy search strategy by maintaining a fixed number of the most promising partial hypotheses, rather than committing to the single best candidate.
By exploring multiple alternatives simultaneously, beam search balances search efficiency and output quality, enabling models to capture better global structures without incurring the computational cost of exhaustive search.

\textbf{Explainable AI and Agentic AI in legal informatics.}
Explainability is a central concern in legal AI, given judicial demands for transparent and interpretable decisions~\cite{XAIlegal2023}. Agent-based architectures enhance transparency and accountability in legal reasoning~\cite{Boella2007}, supporting explainable decision-making in line with legal requirements for fairness.

\section*{The pipeline}
\begin{figure}[htbp] 
    \centering
    \includegraphics[width=0.75\linewidth]{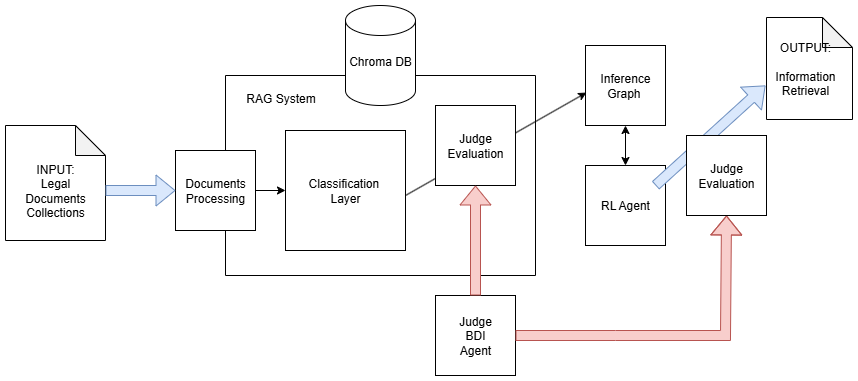} 
    \caption{Illustration of the pipeline. The input legal document is first processed by a classification layer (RAG with Chroma DB embeddings) using MITRE labels. A BDI agent evaluates the classification, and the results are used to construct an inference graph. An RL agent with beam search performs information retrieval, which is further assessed by the Judge agent. Light-blue arrows indicate data flow, while red highlights the Judge agent’s evaluations.}
    \label{fig:my_arch_diagram}
\end{figure}

Our approach is summarized in Figure~\ref{fig:my_arch_diagram}.

\textbf{Dataset Webscraping} The CEPS-Zenner PDF~\cite{zenner2025dataset} was processed to extract EUR-Lex legal document links using PyMuPDF. The pipeline automatically downloads language-specific PDFs via HTML parsing with BeautifulSoup, organizes them in a structured directory, and logs metadata such as CELEX numbers, page references, and file sizes, enabling reproducible multilingual access to the dataset.

\textbf{RAG System} Our RAG system employs a rule-based classifier that leverages approximately 15–20 curated keywords per MITRE ATT\&CK technique. We selected univocal terms for each language, minimizing redundancy with English expressions. While black-box approaches can be highly effective~\cite{PetersBexPrakken2025}, we prioritized transparency. This design not only enhances interpretability and explainability across languages but also produces a structured knowledge graph, which serves as the foundation for subsequent reasoning.

\textbf{Judge Agent} The agent uses a BDI (Belief-Desire-Intention) architecture to autonomously evaluate this multi-agent system. While mindful of the limitations of using an LLM as a judge, it balances these constraints by fully logging beliefs, goals, and reasoning steps, ensuring a transparent, contestable decision trail that supports explainability and accountability.

\textbf{RL Agent} The agent frames the retrieval task as a sequential decision-making problem, where the state space is defined by the graph nodes and the actions are movements to neighboring nodes. It employs a Policy Network that learns to assess the relevance of paths through the graph in response to a natural language query. The search begins by embedding the query and identifying the most relevant tag nodes as entry points. The agent then performs a beam search, expanding a set of candidate paths at each depth. The reward function for path evaluation is a weighted combination of: 1) the semantic similarity between the query and a node's embedding, 2) the confidence weight of the graph edges traversed, and 3) a diversity bonus for paths encompassing multiple unique MITRE tags. This approach allows the agent to reason beyond simple keyword matching, effectively exploring the graph to retrieve a diverse and highly relevant set of legal cases based on the underlying technical tactics and procedures described in the query.
\section*{Initial results}
Current benchmarks inadequately address tasks combining cybersecurity and legal reasoning; we evaluate our pipeline through separate classification and retrieval stages. Evaluation combines qualitative feedback from an \emph{LLM-as-a-Judge}~\cite{zheng2023judgelm} with quantitative analysis on a \emph{golden truth} dataset, built from a cleaned subset of \emph{CyberMetric 10000}~\cite{tihanyi2024cybermetricbenchmarkdatasetbased} and expanded through synthetic augmentation (30:70 real-to-synthetic ratio). A sample of the synthetic data was reviewed by domain experts to ensure plausibility and domain relevance. For the modeling component, we employ as a LLM in our experiments \emph{DeepSeek R1 (distilled with Qwen) 7B parameters}~\cite{deepseekai2025deepseekr1incentivizingreasoningcapability} as an initial, resource-efficient baseline. enriched by synthetic augmentation~\cite{borji2023synthetic,xu2023_ml_synthetic_data_review}. This hybrid approach, further supported by a semi-synthetic cyber dataset, aligns with recent work on evaluation methodologies in AI and law~\cite{surden2019artificial,sanchi2024hybrid}.

\textbf{Classification Results.} On CEPS-Zenner (1,351 samples, 15 labels, 10 languages), the LLM-as-a-Judge evaluation achieved 0.84 accuracy with balanced precision and recall of 0.82, performing well on frequent techniques (e.g., Phishing, Remote Services).
On the Hybrid Cyber 10,000 benchmark (2,940 samples), overall accuracy dropped to 0.455 (macro precision 0.791, recall 0.292, F1 0.397), with strong performance on English inputs (0.834 accuracy) but moderate results on non-English queries, indicating expected cross-lingual generalization, likely domain related~\cite{ranade2018usingdeepneuralnetworks}.
For example, EU regulation CELEX\_32023R2841\_EN was mapped to three ATT\&CK techniques: T1190 (Exploit Public-Facing Application), T1021 (Remote Services), and T1134 (Access Token Manipulation), demonstrating extraction of relevant technical threats from legal text.

\textbf{Retrieval Results.} Using our full pipeline, the LLM-as-a-Judge evaluated retrieval on 255 queries with a context built from CEPS-Zenner documents (20,400 qrels). The system achieved P@1 0.494, P@3 0.431, P@5 0.321, P@10 0.163, recall@10 0.020, F1@10 0.036, MAP 0.019, and MRR 0.552. In a separate evaluation, the retrieval agent was tested on a hybrid-generated context of similar size for the same queries, yielding essentially identical precision and recall values. These results indicate that the agent reliably retrieves top-ranked documents regardless of context source, though overall coverage beyond the top ranks remains limited.

\section*{Conclusions and Future Work}
This study provides initial evidence for the feasibility of combining symbolic reasoning with neural retrieval in a multilingual cyber-legal context, while highlighting areas for improvement. Evaluation used a dual complementary approach: an LLM agent as a judge for scalable assessment and a curated manual evaluation to mitigate reliance on black-box models. Domain-specific comparisons remain for fully legal and cybersecurity benchmarks, with the hybridized dataset partially filling this gap.
On CEPS-Zenner, the rule-based classifier achieved strong aggregate performance, though without language-specific breakdowns. Retrieval showed promising top-rank precision, but recall and MAP remain limited, reflecting effectiveness on directly relevant obligations while broader coverage is still challenging. Results on the hybrid benchmark were weaker, highlighting cross-domain generalization challenges that merit further study. Future work will explore hybrid neural-symbolic embeddings, richer multilingual resources, and human-in-the-loop evaluation with legal and cybersecurity experts. We also plan to expand evaluation using the COLIEE dataset~\cite{rabelo2022coliee}, implement question-answering for comparison with Cybersecurity1000, align the knowledge graph with semantic standards like Akoma Ntoso, and investigate applications in compliance and audit contexts.
\bibliographystyle{vancouver.bst}
\bibliography{biblio}

\end{document}